\documentclass[letterpaper, 10pt, conference]{ieeeconf} 
\usepackage{amsmath,amssymb,amsfonts}
\usepackage{graphicx}
\usepackage{textcomp}
\usepackage{graphics} 
\usepackage{epsfig} 
\usepackage{subfigure}
\usepackage{amsmath} 
\usepackage{amssymb}  
\usepackage[outdir=./figs/]{epstopdf}
\usepackage{lipsum}
\usepackage{lipsum,amsmath,multicol}
\usepackage{float}
\usepackage{algpseudocode}
\usepackage{wrapfig}
\usepackage{sidecap}

\usepackage{amsthm}

\usepackage{comment}
\usepackage{array}
\usepackage{glossaries}
\usepackage{breqn}
\usepackage{booktabs}
\newcolumntype{L}[1]{>{\raggedright\let\newline\\\arraybackslash\hspace{0pt}}m{#1}}
\newcolumntype{C}[1]{>{\centering\let\newline\\\arraybackslash\hspace{0pt}}m{#1}}
\newcolumntype{R}[1]{>{\raggedleft\let\newline\\\arraybackslash\hspace{0pt}}m{#1}}
\usepackage{sidecap}
\usepackage{url}
\usepackage[T1]{fontenc}

\usepackage{hyperref}

\usepackage{tabularx}
\usepackage{multicol, blindtext}
\usepackage{amsmath}
\usepackage{multirow}
\usepackage[dvipsnames]{xcolor}
\usepackage{xcolor}

\usepackage{optidef}
\definecolor{my_green}{RGB}{144, 238, 144}
\usepackage[linesnumbered,ruled,vlined]{algorithm2e}
\SetAlFnt{\small}
\SetAlCapFnt{\small}
\usepackage{color, colortbl}

\newcommand\blfootnote[1]{%
  \begingroup
  \renewcommand\thefootnote{}\footnote{#1}%
  \addtocounter{footnote}{-1}%
  \endgroup
}

\begin{document}

\title{\LARGE \bf Differentiable-Optimization Based Neural Policy for Occlusion-Aware Target Tracking }
\author{Houman Masnavi, Arun Kumar Singh, and Farrokh Janabi-Sharifi}
\maketitle

\blfootnote{Houman Masnavi and Farrokh Janabi-Sharfi are with the Toronto Metropolitan University, Canada and Arun Kumar Singh is with the University of Tartu, Estonia. GitHub repository: https://github.com/Houman-HM/diff_opt_tracking}

\begin{abstract} 
Tracking a target in cluttered and dynamic environments is challenging but forms a core component in applications like aerial cinematography. The obstacles in the environment not only pose collision risk but can also occlude the target from the field-of-view of the robot. Moreover, the target future trajectory may be unknown and only its current state can be estimated.  In this paper, we propose a learned probabilistic neural policy for safe, occlusion-free target tracking.

The core novelty of our work stems from the structure of our policy network that combines generative modeling based on Conditional Variational Autoencoder (CVAE) with differentiable optimization layers. The role of the CVAE is to provide a base trajectory distribution which is then projected onto a learned feasible set through the optimization layer. Furthermore, both the weights of the CVAE network and the parameters of the differentiable optimization can be learned in an end-to-end fashion through demonstration trajectories.

We improve the state-of-the-art (SOTA) in the following respects. We show that our learned policy outperforms existing SOTA in terms of occlusion/collision avoidance capabilities and computation time. Second, we present an extensive ablation showing how different components of our learning pipeline contribute to the overall tracking task. We also demonstrate the real-time performance of our approach on resource-constrained hardware such as NVIDIA Jetson TX2. Finally, our learned policy can also be viewed as a reactive planner for navigation in highly cluttered environments.

\end{abstract}

\section{Introduction}
\label{sec:introduction}
The ability to track a target while avoiding occlusion and collision from the environment forms a key building block in many applications such as cinematography \cite{drone_cinema_1}, cooperative navigation \cite{martin_saska_paper}, \cite{vacna_paper}, robot-assistant \cite{nishimura2006study} in hospitals, and warehouses. There are two core challenges towards computing an optimal tracking trajectory. First, the future trajectory of the target is invariably unknown and the robot only has the estimates of its instantaneous position and velocity. Secondly, trajectory smoothness, collision, and occlusion often conflict with each other \cite{ral_vis_aware}, \cite{fei_gao_visibility}, \cite{auto_chaser_1} creating local minima traps for the optimization process.

In this paper, we propose a learning-based approach for addressing the core challenges described above and enabling safe, agile, and occlusion-free target tracking in arbitrary environments. 
Specifically, we learn a policy that maps environment conditions (point clouds) along with robot and target states to a distribution of optimal trajectories. Subsequently, the robot ranks the sampled trajectories from the learned distribution in terms of smoothness, collision and occlusion cost to ascertain the best trajectory for execution. Our approach is light-weight and can deliver real-time performance even on resource-constrained hardware. As a result, it can quickly adapt to sharp changes in target's motion and deliver high fidelity tracking with just linear prediction of target trajectories. The main contributions of the work are summarized below:

\begin{figure}[!t]
\includegraphics[scale=0.29]{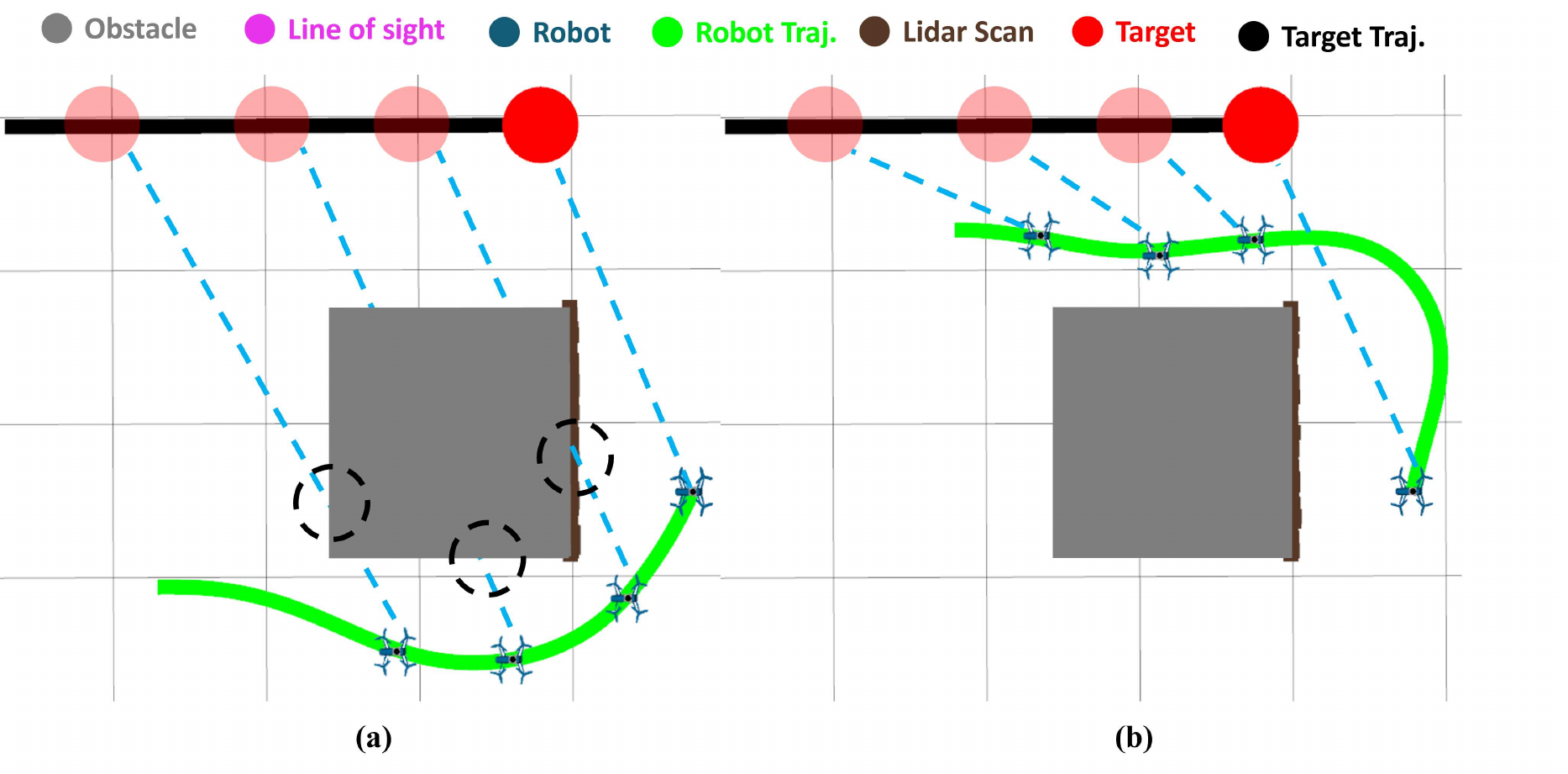}
\vspace{-0.7cm}
\caption{ (a) and (b) show top-down views of regular target tracking vs. occlusion-aware target tracking, respectively. When there is no occlusion requirement (a), the robot tracks the target while avoiding the obstacle from the bottom. However, when the occlusion requirement is added, it forces the robot to avoid the obstacle from the top to keep its line of sight unobstructed.}
  \label{notation_figure}
  \vspace{-0.7cm}
\end{figure}

\noindent \textbf{Algorithmic:} We propose a novel probabilistic policy that consists of a Conditional Variational Autoencoder (CVAE) \cite{kingma_cvae} embedded with a differentiable optimization layer. The CVAE provides a base distribution of trajectories while the optimization layer acts as a safety filter that projects the sampled trajectories onto a paramertized feasible set. We propose a customized end-to-end training for simultaneously learning the weights of the CVAE and the parameters of the optimization layers through demonstration of optimal trajectories. In particular, we reformulate the constraints of the embedded optimization in a form that allows for efficient backpropagation through the optimization steps during the end-to-end training.

\noindent \textbf{State-of-the-art Performance:} We compare our approach extensively with two state-of-the-art (SOTA) baselines \cite{ral_vis_aware}, \cite{auto_chaser_1} and show massive improvement in the ability to perform occlusion and collision-free tracking, especially at high target speeds. We also compare against a conventional behaviour cloning (BC) approach to highlight the importance of embedding a differentiable optimization layer within the CVAE architecture. Moreover, we also demonstrate how learning the parameters of optimization layers during end-to-end training leads to better tracking performance as compared to hand-crafting them based on some nominal values. Finally, we validate real-time inferencing capabilities of our approach on embedded devices such as NVIDIA Jetson TX2.

\section{Problem Formulation}

\noindent \textbf{Symbols and Notations:} Regular small-case letters denote scalars, whereas their bold font variants signify vectors. Matrices are denoted by bold uppercase letters. The variable $t$ serves as the timestamp for a variable, and the uppercase variant $T$ indicates the transpose of a matrix. The key symbols are outlined in Table \ref{symbols}, with some symbols being defined at their initial occurrence.

\noindent \textbf{Assumptions}

\begin{itemize}

    \item For the ease of exposition, we consider the motion in the 2D, X-Y plane. However, our approach is trivially extendable to 3D setting 

  \item Similar to prior works such as \cite{fast_tracker_1} and \cite{drone_cinema_3}, we assume that the robot has a front facing camera whose orientation control is independent of the translational motion along the $X-Y$ axis.
\end{itemize}

\begin{table}[!t]
\centering
\caption{\scriptsize Important Symbols} \label{symbols}
\vspace{-0.2cm}
\scriptsize
\begin{tabular}{|p{3.3cm}|p{4cm}|}\hline
$\textbf{p} = (x(t), y(t))$ & Position of the robot at time $t$.\\ \hline
$\textbf{p}_r = (x_{r}(t), y_{r}(t))$ & Position of the target at time $t$.\\ \hline
$\textbf{p}_{o, i} = (x_{o, i}(t), y_{o, i}(t)$ & $i^{th}$ LiDAR point at time $t$.\\ \hline
$n, m$ & Batch-size and planning horizon, respectively.\\ \hline
\end{tabular}
\vspace{-0.7cm}
\end{table}

\normalsize
\subsection{Optimization for Target Tracking}
\noindent We can formalize occlusion-aware target tracking as the following optimization problem:
\vspace{-0.1cm}
\begin{subequations}
\begin{align}
    \min \sum_{t = t_0}^{t=t_m} w_1(\ddot{\mathbf{p}}(t))^2 + w_2f_{occ}(\mathbf{p}(t), \mathbf{p}_r(t)) \label{cost}\\
    (\mathbf{p}(t_0), \dot{\mathbf{p}}(t_0), \ddot{\mathbf{p}}(t_0)) = \mathbf{b}_{0} \label{uav_init_boundary_condition} \\
    ({\mathbf{p}}(t_m), \dot{\mathbf{p}}(t_m)) = \mathbf{b}_{f} \label{uav_final_boundary_condition}\\
    f_{v}(\dot{\mathbf{p}}(t))\leq v_{max}, \forall t \label{uav_v_bounds}\\
    f_{a}(\ddot{\mathbf{p}}(t))\leq a_{max}, \forall t \label{uav_acc_bounds} \\
    \underline{s}_{los}(t) \leq \left \Vert \begin{bmatrix} \mathbf{p}(t)-\mathbf{p}_r(t)
    \end{bmatrix} \right \Vert_2\leq \overline{s}_{los}(t), \forall t \label{uav_tracking}\\
    \Vert \mathbf{p}(t)-\mathbf{p}_{o,i} \Vert_2^2 -l^2 \leq 0, \forall t \label{coll_avoid}
\end{align}    
\end{subequations}

\normalsize
\vspace{-0.5cm}
\small
\begin{align}
    f_{v} = \Vert\dot{\mathbf{p}}(t)\Vert_2, f_{a} = \Vert\ddot{\mathbf{p}}(t)\Vert_2
    \label{uav_acc_vel_const_functions}
\end{align}

\normalsize
\noindent The vectors $\mathbf{p} = [x(t),\:y(t)]^T $, and $\mathbf{p}_r = [x_{r}(t),\:y_{r}(t)]^T $ represent the 2D positions of the robot and the target respectively. The first term in the cost function \eqref{cost} minimizes the sum of the acceleration magnitude across the robot trajectory. The second term minimizes the occlusion between the robot and the target computed at discrete time instants in the planning horizon. Several occlusion models are possible, and in this work, $f_{occ}$ is a neural network, trained to predict occlusion values based on point cloud \cite{visibility_aware}. However, our approach is agnostic to the nature of the cost function. The constants $w_1 - w_2$ are user-defined weights utilized to trade off the relative importance of each cost term. Constraints \eqref{uav_init_boundary_condition}  - \eqref{uav_final_boundary_condition} define the initial and final boundary conditions on the trajectory. Therefore, the vectors $\textbf{b}_{0},\textbf{b}_{f}$ are simply stacking of robot's initial and final positions, velocities, and accelerations. The inequality constraints \eqref{uav_v_bounds}-\eqref{uav_acc_bounds} maintain the norm of velocities and accelerations across the trajectory within specified bounds. The definitions of velocity and acceleration constraints are given in \eqref{uav_acc_vel_const_functions}. The inequality \eqref{uav_tracking} ensures that the robot and the target stay within certain bounds ($\underline{s}_{los}(t), \overline{s}_{los}(t)$) from each other. The tracking setup is depicted in Figure \ref{notation_figure}. The last set of inequality \eqref{coll_avoid} models the collision avoidance constraints, wherein we treat $i^{th}$ LiDAR point-cloud as a small disk-shaped obstacle with center $\mathbf{p}_{o, i}$ and radius $l$.

\subsubsection{Polynomial Parameterization} We parameterize the position level trajectory of the robot in terms of polynomials in the following form:
\small
\begin{align}
    \begin{bmatrix}
        x(t_1), \dots, x(t_m) 
    \end{bmatrix} = \mathbf{W}\textbf{c}_{x},
     \begin{bmatrix}
        y(t_1), \dots, y(t_m) 
    \end{bmatrix} = \mathbf{W}\textbf{c}_{y},
    \label{param}
\end{align}
\normalsize

\noindent where, $\mathbf{W}$ is a matrix formed with time-dependent polynomial basis functions and $\textbf{c}_{x}, \textbf{c}_{y}$ are the coefficients of the polynomial. We can also express the derivatives in terms of $\dot{\mathbf{W}}, \ddot{\mathbf{W}}$.

Using \eqref{param}, the optimization \eqref{cost}-\eqref{uav_tracking} can be compactly represented as :
\vspace{-0.1cm}
\begin{subequations}
    \begin{align}
    \boldsymbol{\xi}^*(\mathbf{q})  =     \arg \min_{\boldsymbol{\xi}} c(\boldsymbol{\xi}) \label{cost_reform}  \\
        \mathbf{A}\boldsymbol{\xi} = \mathbf{b}(\mathbf{q}) \label{eq_reform} \\
        \mathbf{g}(\boldsymbol{\xi}, \mathbf{q}) \leq 0 \label{ineq_reform}
    \end{align}
\end{subequations}

\noindent where $\mathbf{q} = (\underline{s}_{los}(t), \overline{s}_{los}(t), \mathbf{b}_f) $. The equality constraints \eqref{eq_reform} are the boundary conditions of \eqref{uav_init_boundary_condition}-\eqref{uav_final_boundary_condition}, while the inequality constraints \eqref{uav_v_bounds}-\eqref{uav_tracking} are rolled into \eqref{ineq_reform}. We discusses their algebraic forms later in the paper. However, the key thing to note here is that the optimal solution is parameterized by the parameter $\mathbf{q}$ consisting of min/max separation between the robot and the target and the terminal position/velocity. 
As can be seen, the parameter $\mathbf{q}$ characterizes the constraint sets and thus, has a significant effect on the tracking efficiency. 

In the next section, we present a data-driven approach for solving \eqref{cost_reform}-\eqref{ineq_reform}, where we simultaneously obtain not only the optimal solution $\boldsymbol{\xi}^*$ but also the associated parameter $\mathbf{q}$.

\normalsize
\section{Main Algorithmic Results}

\begin{figure}[!t]
    \centering
    \includegraphics[scale=0.405]{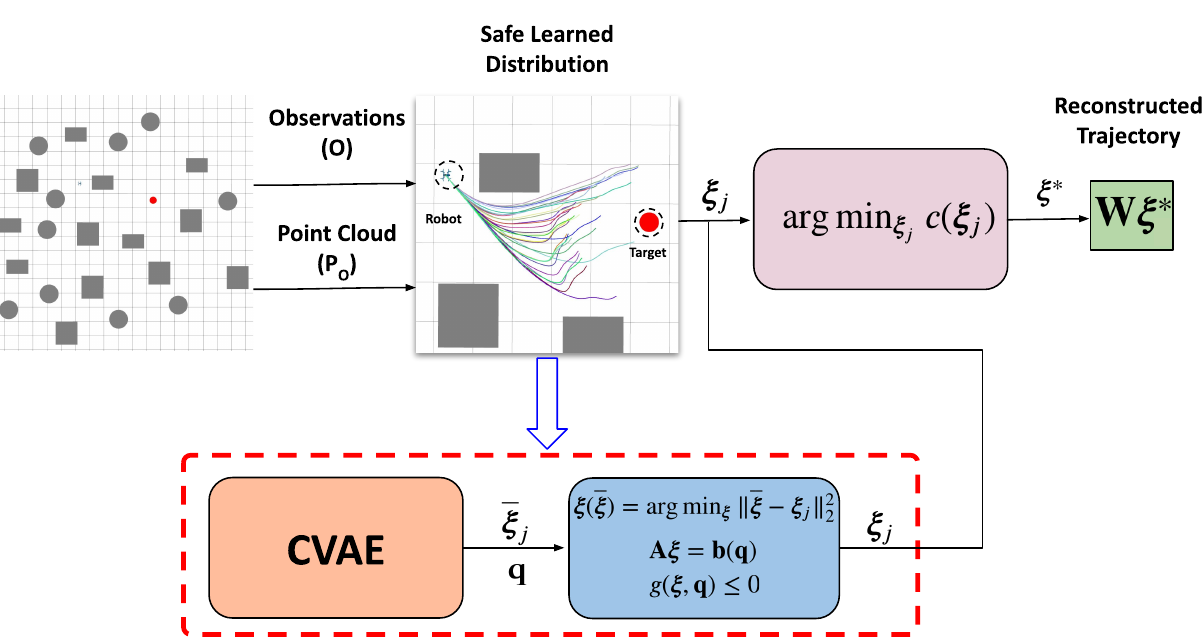}
    \vspace{-0.2cm}
    \caption{Our learning-based approach for solving \eqref{cost_reform}-\eqref{ineq_reform} that relies on sampling trajectory coefficients $\boldsymbol{\xi}_j$ from a learned distribution conditioned on the observations (point clouds, states). The samples $\boldsymbol{\xi}_j$ are sorted based on their associated cost and the one with the least value is selected as the optimal solution. To ensure that sampled $\boldsymbol{\xi}_j$ lead to safe and kinematically feasible trajectories, the learned distribution is structured in the form of a CVAE augmented with a differentiable projection optimizer.}
    \label{fig:pipeline_overview}
    \vspace{-0.5cm}
\end{figure}

Fig. \ref{fig:pipeline_overview} presents our overall learning-based approach for approximately solving optimization \eqref{cost_reform}-\eqref{ineq_reform}. We sample trajectory coefficient  $\boldsymbol{\xi}_j$ from a learned distribution and evaluate the cost $c(\boldsymbol{\xi}_j)$ over these samples. Subsequently, the optimal solution is defined as the sample leading to the minimum cost. Although relatively simple, such sampling-based approach has proved tremendously useful in generating highly-agile and reactive behaviors, especially when the sampling is done from a learned policy \cite{howell2022predictive}. The core novelty of our approach lies in ensuring that the samples drawn from the learned distribution are safe (collision-free) and satisfy the kinematic bounds. As shown, the learned distribution has two components namely a CVAE and a projection optimizer. The former provides a base distribution from which nominal samples $\overline{\boldsymbol{\xi}}_j$ are drawn, which are then projected onto the $j^{th}$ learned feasible set characterized by $\mathbf{q}_j$. 

In the following subsections, we present further details on end-to-end training of our projection optimizer embedded CVAE and how we ensure computational tractability of backpropagating through the optimization layer.

\subsection{CVAE with Differentiable Optimization Layers}

\noindent Assume that we are given a dataset of expert optimal trajectories $\boldsymbol{\tau}_e$ for given observations of the robot and the target state $\mathbf{o}$  and vector $\mathbf{p}_o$ which is obtained by stacking all the point-cloud/LiDAR scans $\mathbf{p}_{o,i}$. Our aim in this section is to learn the distribution of optimal trajectories, $\boldsymbol{\pi}$ conditioned on $(\mathbf{o}, \mathbf{p}_{o})$. To this end, we model $\boldsymbol{\pi}$ in the form of a novel CVAE and train it in end-to-end manner with a behavior cloning loss on  $\boldsymbol{\tau}_e$.

\begin{figure}[!t]
    \centering
    \includegraphics[scale=0.43]{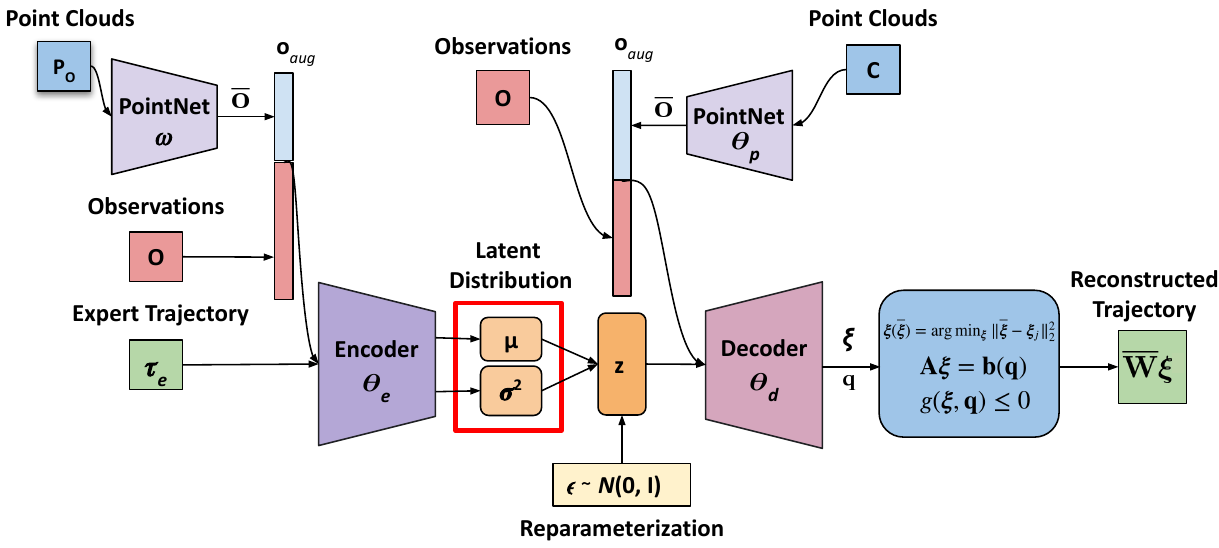}
    \vspace{-0.6cm}
    \caption{Proposed CVAE architecture augmented with a differentiable optimization layer. We use PointNet to encode point-clouds to some latent features as a part of the conditioning of the CVAE.}
    \label{fig:cvae_pipeline}
    \vspace{-0.6cm}
\end{figure}

Our CVAE architecture is shown in Fig. \ref{fig:cvae_pipeline}. It consists of an encoder-decoder architecture augmented with a PointNet \cite{Qi_2017_CVPR} that extracts features $\overline{\mathbf{o}}$ from the point-cloud $\mathbf{p}_{o}$. The learnable parameters of PointNet are depicted by $\boldsymbol{\theta}_p$. We stack the features from PointNet and state observations to create an augmented vector $\textbf{o}_{aug} = (\textbf{o}, \overline{\textbf{o}})$. This is then fed to the encoder of the CVAE modeled in the form of a multi-layer perceptron (MLP), along with  $\boldsymbol{\tau}_e$. The encoder with learnable weights $\boldsymbol{\theta}_e$ maps $(\mathbf{o}_{aug}, \boldsymbol{\tau}_e)$ to a latent distribution $\mathbf{z} \sim \mathcal{N} (\boldsymbol{\mu}_{\boldsymbol{\theta}_e}, diag(\boldsymbol{\sigma^2}_{\boldsymbol{\theta}_e}))$.

The decoder model is also an MLP with parameters $\boldsymbol{\theta}_d$ augmented with a differentiable projection optimizer. It takes in $\textbf{o}_{aug}$ and maps $\textbf{z}$ to the output distribution $\boldsymbol{\pi}(\overline{\boldsymbol{\xi}}|\textbf{z}, \textbf{o}_{aug})$, which provides a base distribution for sampling optimal trajectory polynomial coefficients $\overline{\boldsymbol{\xi}}$. However, the resulting trajectories are unlikely to satisfy the kinematic and collision avoidance constraints. Therefore, $\overline{\boldsymbol{\xi}}$ is passed through the projection optimization layer to obtain $\boldsymbol{\xi}$ that leads to feasible and safe trajectories.
Moreover, as shown in Fig. \ref{fig:cvae_pipeline}, the decoder MLP not only outputs $\overline{\boldsymbol{\xi}}$ but also the parameters $\mathbf{q}$ defining the constraints and $({^0}\boldsymbol{\lambda}, {^0}\boldsymbol{\xi})$ (to be defined later) that act as the initialization of the projection optimizer. Essentially, the aim of the decoder is to not only learn an effective base distribution but also to shape the projection optimizer and accelerate its convergence. 

The encoder, decoder, and PointNet are all trained in an end-to-end manner. In a typical CVAE, the training process reduces to an unconstrained optimization over the weights $\boldsymbol{\theta} = (\boldsymbol{\theta}_p, \boldsymbol{\theta}_e, \boldsymbol{\theta}_d,)$. In sharp contrast, the presence of the projection layer converts the training process into a bi-level optimization problem, as shown in \eqref{loss}-\eqref{ineq_reform_2}. The first two terms in the loss function \eqref{loss_2} are typical of any CVAE and respectively minimize the reconstruction loss and difference of the latent distribution from a normal distribution. In \eqref{loss}, $\overline{\mathbf{W}}$ is formed by diagonally stacking $\mathbf{W}$. The third term in \eqref{loss_2} is unique to our training pipeline and encourages the projection optimizer to converge to solutions with low inequality constraint residuals. This in turn,  ensures safety and kinematic feasibility of the trajectories resulting from $\boldsymbol{\xi}$. The $\beta$ in \eqref{loss_2} controls the relative importance of the KL loss with respect to other terms and is a tuneable parameter.

\vspace{-0.4cm}

\small
\begin{subequations}
\begin{align}
     \min_{(\boldsymbol{\theta})} \mathcal{L}(\boldsymbol{\xi} ) \label{loss}\\ 
    \text{such that}\hspace{0.3cm} \boldsymbol{\xi} = \arg\min_{\boldsymbol{\xi}} \Vert \boldsymbol{\xi}-\boldsymbol{\overline{\xi}}(\boldsymbol{\theta} ) \Vert_2^2  \label{cost_reform_2} \\
        \mathbf{A}\boldsymbol{\xi} = \mathbf{b}(\mathbf{q}(\boldsymbol{\theta} )) \label{eq_reform_2} \\
        \mathbf{g}(\boldsymbol{\xi}, \mathbf{q}(\boldsymbol{\theta}) \leq 0 \label{ineq_reform_2}
    \end{align}    
\end{subequations}
\vspace{-0.7cm}
\normalsize
\small
\begin{dmath}
    \mathcal{L}(\boldsymbol{\xi}) =  \hspace{-.11cm}\min\sum \left\Vert \overline{\mathbf{W}}\boldsymbol{\xi} - \boldsymbol{\tau}_e \right \Vert_2^2 \newline
    + \beta \, {D}_{\mathbf{KL}}[\mathcal{N} (\boldsymbol{\mu}_{\theta_e}, diag(\boldsymbol{\sigma^2}_{\theta_e})) | \,\mathcal{N}(\textbf{0}, \textbf{I}  )  ]+\Vert \max(\mathbf{0}, \mathbf{g}( \boldsymbol{\xi}, \mathbf{q} (\boldsymbol{\theta}) ))\Vert_2^2, \label{loss_2}
\end{dmath}
\vspace{-0.35cm}
\normalsize

\subsection{Differentiation Through the Optimization Layer}
\noindent Minimization of \eqref{loss} requires us to compute $\nabla_{\boldsymbol{\theta}}\mathcal{L}$ which can be defined in terms of chain rule in the following manner.
\vspace{-0.1cm}
\small
\begin{align}
    \nabla_{\boldsymbol{\theta}}\mathcal{L}(\boldsymbol{\xi} ) = \nabla_{\boldsymbol{\xi}}\mathcal{L}(\boldsymbol{\xi})\colorbox{my_green}{$\nabla_{\boldsymbol{\theta}}{\boldsymbol{\xi}}(\boldsymbol{\theta})$}
    \label{chain_rule}
\end{align}
\normalsize

\begin{figure*}
    \centering
    \includegraphics[scale = 0.3]{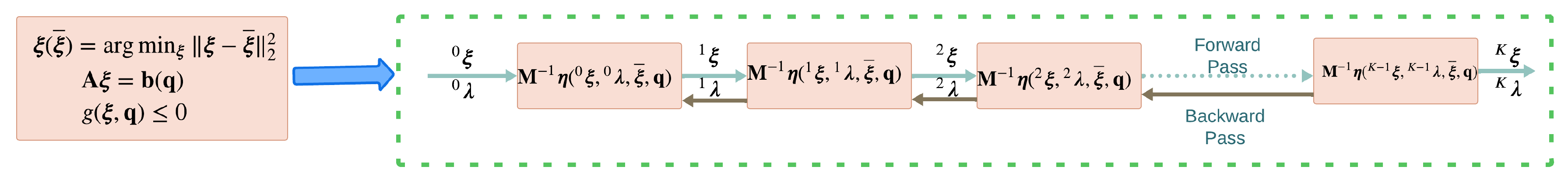}
    \vspace{-0.7cm}
    \caption{The unrolled structure of our differentiable projection optimizer. It includes solving a sequence of linear systems. We can backpropagate through the optimizer iterations to compute how ${^K}\boldsymbol{\xi}$ would change with respect to the CVAE decoder output $\overline{\boldsymbol{\xi}}, {^0}\boldsymbol{\xi}, {^0}\boldsymbol{\lambda}, \mathbf{q}$.}
    \label{fig:proj_iteration}
    \vspace{-0.5cm}
\end{figure*}

\noindent Computing the highlighted term on the right-hand-side of \eqref{chain_rule} requires us to differentiate the solution of the inner optimization \eqref{cost_reform_2}-\eqref{ineq_reform_2} with respect to its input parameter $\boldsymbol{\theta}$. There are two ways to perform this operation namely, implicit differentiation and algorithm unrolling \cite{pineda2022theseus}. The former is based on differentiating the optimality conditions and thus cannot be used to learn a good initialization\footnote{The initial guess used to start an optimizer does not explicitly appear in the optimality conditions. Thus, it cannot be learned through implicit differentiation.}. On the other hand, algorithm unrolling does not have such limitations but requires each step of the optimization solver to be differentiable. To this end, we propose custom differentiable optimization routine for solving \eqref{cost_reform_2}-\eqref{ineq_reform_2}. We present a detailed analysis in the Appendix \ref{append} but provide the higher-level insights here.

The numerical steps of our projection optimizer can be reduced to a fixed-point operation of the following form, wherein $k$ represents the iteration index.
\begin{align}
    ({^{k+1}}\boldsymbol{\xi}), {^{k+1}}\boldsymbol{\lambda} = \mathbf{M}^{-1}\boldsymbol{\eta}({^k}\boldsymbol{\xi},~  {^{k}}\boldsymbol{\lambda}, ~\overline{\boldsymbol{\xi}},~ \mathbf{q})
    \label{fixed_point_form}
\end{align}

\noindent The matrix $\mathbf{M}$ and the vector $\boldsymbol{\eta}$ are derived in the Appendix \ref{append}. However, we highlight a few points about these entities here. First, the vector $\boldsymbol{\eta}$ has a closed-form expression and depends on the previous iteration solution ${^k}\boldsymbol{\xi}$ as well as the output of the CVAE decoder $(\overline{\boldsymbol{\xi}}, \mathbf{q})$. Second, we show in the Appendix \ref{append} that the matrix $\mathbf{M}$ does not have any learnable parameters and is also independent of the ${^k}\boldsymbol{\xi}$. As a result, its inverse needs to be computed only once before the training process starts. This in turn, substantially improves the numerical stability as well as the computational speed of our training pipeline.

As shown in Fig. \ref{fig:proj_iteration}, we run our projection optimizer for a fixed number of iterations $K$ and approximate $\nabla_{\boldsymbol{\theta}}\boldsymbol{\xi}$ by tracing the gradient through the fixed-point iterations. Typically, the unrolled-chain length $K$ needs to be small to avoid vanishing/exploding gradient issues and to keep the memory footprint of the training pipeline small. 

\section{Connections to Prior Works}
\noindent \textbf{Model based approaches:} Occlusion-aware target tracking is commonly addressed through a combination of graph-search and trajectory optimization approaches \cite{fei_gao_visibility}, \cite{auto_chaser_1}, \cite{ji2022elastic}. These works use the signed-distance field based on prior computed maps to compute collision and occlusion costs. In contrast, our prior work \cite{ral_vis_aware} used a learned model to directly predict occlusions over point-clouds. This in turn, allowed target tracking in dynamic environments. Our current work inherits the same advantages from \cite{ral_vis_aware}.  It is also possible to estimate occlusion in the image frame as shown in \cite{penin2018vision}. A key limitation of model based approaches is that several parameters like minimum and maximum tracking distance need to be manually tuned. Moreover, prior demonstrations cannot be leveraged for improving the target tracking performance or accelerate the computational performance of the planners. Our proposed work addresses these limitations.

\noindent\textbf{Differentiable Optimization Based Imitation Learning:} One of the fundamental challenges in neural network based planning is that the predicted trajectories may not satisfy the safety/collision-avoidance or kinematic constraints. Recently, differentiable optimization layers have emerged as a potential solution \cite{amos2018differentiable}, \cite{xiao2023barriernet} towards this problem. The core idea is to embed optimization solvers as layers into the neural network pipeline. The backpropagation traces the gradient through the solvers and provides corrective feedback to the neural-network predictions during training. However, most of the success of differentiable optimization embedded learning has come while working with convex problems. In sharp contrast, our proposed work demands embedding a more challenging non-convex solver, \eqref{cost_reform_2}-\eqref{ineq_reform_2}. Moreover, the computational structure highlighted in Appendix \ref{append} allows for efficient backpropagation through the optimization layer.

\noindent \textbf{Contribution over Author's Prior Work:} The current work improves our recent efforts in differentiable optimization layers based end-to-end learning \cite{shrestha2023end} for autonomous driving. Specifically, unlike \cite{shrestha2023end}, the proposed differentiable optimization layer has learnable parameters which define the constraint set. As we show in Section \ref{ablation}, the learned constrained set outperforms that obtained by manually tuning the parameters of the constraint set. 

\section{Validation and Benchmarking}
\noindent This section aims to answer the following research questions:

\begin{itemize}
    \item \textbf{Q1:} How does our learning-based approach compare to existing SOTA approaches based on model predictive control and graph search?
    \item \textbf{Q2:} What is the performance gain achieved by augmenting CVAE with a differentiable optimization layer?
    \item \textbf{Q3:} What is the advantage of learning the parameters $\mathbf{q}$ of the constraints in the projection optimizer vis-a-vis fixing them based on heuristics?
\end{itemize}

\subsection{Implementation Details} 
\noindent We implement our training pipeline shown in Fig. \ref{fig:cvae_pipeline} in PyTorch \cite{pytorch}. However, for faster inferencing, we re-implement the projection optimizer in JAX \cite{jax2018github}. The simulation pipeline was built on top of Robot Operating System (ROS) \cite{ros} and the Gazebo physics simulator. All the benchmarks were mainly run on AORUS Laptop Intel core $i7-11800H$ with NVIDIA RTX 3080 and NVIDIA Jetson TX2.  
\subsection{CVAE Training} 
\noindent During training, the inputs to the CVAE are the point-clouds and the states (position, velocity) of the robot as well as the target. The robot's position is always at the origin. Thus, the point-clouds and target states are shifted accordingly. The outputs of the CVAE are the trajectory coefficients for the initialization of the projection optimizer along with its  constraint parameters. The predicted coefficients produce a trajectory over a 5-second horizon. The details of the PointNet, Encoder, and Decoder MLP of the CVAE are provided in the accompanying video. The demonstration of the optimal trajectory was obtained through the sampling-based approach presented in \cite{ral_vis_aware}, run in offline mode with a large sample size of 5000. Thus, in essence our learning-based approach aims to distill the knowledge of a more resourced and complicated optimizer into a computationally efficient policy network.

\begin{figure*}[!h]
\includegraphics[scale=0.55]{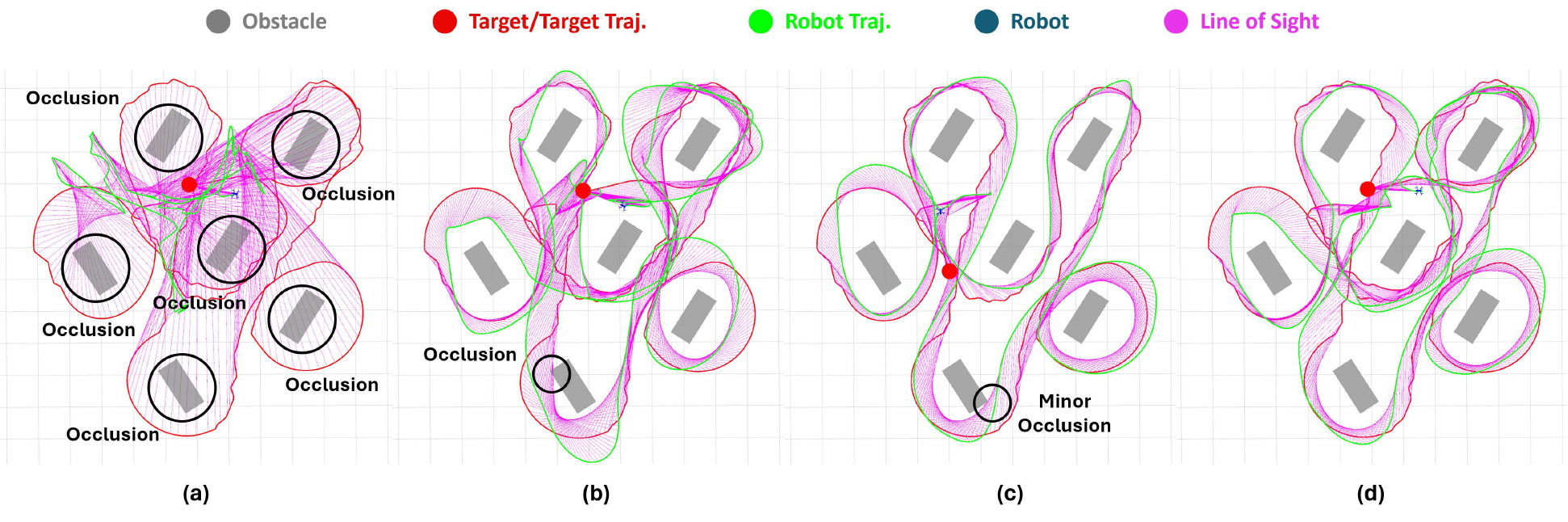}
\vspace{-0.8cm}
\caption{ Target tracking results for \textbf{AutoChaser} \cite{auto_chaser_1} (a), \textbf{Proj-CEM} \cite{ral_vis_aware} (b), conventional Behavior cloning (\textbf{BC}) (c) and our proposed approach (d), respectively. The environment consists of 6 obstacles and the target is moving with the max speed of 1m/s. AutoChaser gets occluded around each obstacle, while \textbf{Proj-CEM} \cite{ral_vis_aware} and the conventional \textbf{BC} perform much better and only has minor occlusion around one of the obstacles. Our proposed method finishes the task successfully without any occlusion of the target throughout the whole run.}
  \label{1m_comparison_figure}
  \vspace{-0.5cm}
\end{figure*}


\subsection{Baselines}
\noindent\textbf{AutoChaser} \cite{auto_chaser_1}: We compare our learning-based approach with \cite{auto_chaser_1} that combines graph-search with trajectory optimization. The approach of \cite{auto_chaser_1} relies on knowing the intermediate target goal points to predict its future trajectory. Thus, for a fair comparison, we provided a dense set of intermediate target goal points to \cite{auto_chaser_1}. It should be noted that all the other baselines including our approach have access to only the current, instantaneous position and velocity of the target. We also create a variant of \cite{auto_chaser_1}, where the target future intermediate goals are not provided.

\noindent\textbf{Proj-CEM \cite{ral_vis_aware}}: The approach of \cite{ral_vis_aware} can be considered to be the current SOTA that combines the Cross-Entropy Method (CEM) with convex optimization over a learned occlusion model. This baseline is particularly important as our proposed approach can be seen as learning-enhanced version of it. Specifically, our approach samples from a learned distribution instead of a naive Gaussian used in the CEM of \cite{ral_vis_aware}. Moreover, in contrast to \cite{ral_vis_aware}, our projection optimizer is learned instead of being hand-crafted. Finally, unlike \cite{ral_vis_aware}, our approach does not require multiple refinement of the sampling distribution, which improves its computational performance. 

\noindent\textbf{Behavior Cloning (BC)}: This is a conventional imitation learning approach wherein we just remove the projection optimizer from the CVAE pipeline depicted in Fig. \ref{fig:cvae_pipeline}. Thus, this baseline is designed to highlight the role of the differentiable optimization layers in our approach.

\subsection{Metrics} 

\noindent\textbf{Occlusion Time}: The overall time during which the robot and the target's line of sight (LoS) intersects with an obstacle.

\noindent\textbf{Acceleration}: This measures how quickly the robot needs to change its speed and direction to avoid occlusions/collisions while following the target.

\noindent\textbf{Computation Time}: The time taken to generate a solution trajectory.

\noindent\textbf{Success Rate}: The ratio of occlusion-free runs to the total number of runs in each benchmark.

\subsection{Target Tracking in Static Environments}
\label{static_enironment_section}

\noindent This section evaluates the performance of our approach in environments with static obstacles. We applied our CVAE policy network in a receding horizon fashion to create an implicit feedback control of the robot. We generated 15 distinct target trajectories in three diverse obstacle environments (with target speeds ranging from 0.25 m/s to 2.0 m/s) using tele-operation. These trajectories were subsequently replayed during the simulation. The total duration of the trajectory run in these environments was approximately 2300 seconds. 

\subsubsection{Tracking with a maximum target speed of 1m/s} \leavevmode

 A qualitative representation of the results is depicted in Fig. \ref{1m_comparison_figure}. As it can been seen, \textbf{AutoChaser} \cite{ral_vis_aware} (Fig. \ref{1m_comparison_figure}(a)) gets occluded around each obstacle while the other two baselines (Figs. \ref{1m_comparison_figure}(b) - (c)) perform the task with minor occlusions around one of the obstacles. Our proposed optimizer (Fig. \ref{1m_comparison_figure}(d)), however, successfully finishes the task with no occlusions.
 
Presented in the first half of Table \ref{1m_comp_table} are the quantitative results for this benchmark. As stated earlier, we tested two variants of the \textbf{AutoChaser} \cite{auto_chaser_1}, one that has access to target intermediate waypoints and one that does not. When there is no access to the intermediate waypoints, \textbf{AutoChaser} performs very poorly in terms of the occlusion-time and success-rate. This variant did not manage to have an occlusion free run in any of the trajectory runs. When the intermediate waypoints were provided, the performance improved significantly and the occlusion-time almost halved, but the success-rate achieved was quite low at 0.1. \textbf{Proj-CEM} \cite{ral_vis_aware} and \textbf{BC} achieved better performance by achieving higher success-rate and less occlusion-time. Our approach outperformed all the baselines by successfully finishing task in these scenarios with no collusion of the target. All the baselines used similar acceleration efforts. \textbf{AutoChaser} had the highest computation time of 0.085s. \textbf{BC} achieved better results while taking approximately $84 \%$ less time compared to \textbf{Proj-CEM}. \cite{ral_vis_aware}. Due to the augmentation of the differentiable optimization layer, our optimizer took almost twice the time as that of the \textbf{BC}, however, it achieved superior results in rest of the metrics. It's noteworthy that \textbf{AutoChaser} \cite{auto_chaser_1} requires prior access to the environment map, whereas other baseline methods operate solely on the instantaneous LiDAR data.

\subsubsection{Tracking with a maximum target speed of 2m/s} \leavevmode

\begin{figure*}[!t]
\includegraphics[scale=0.50]{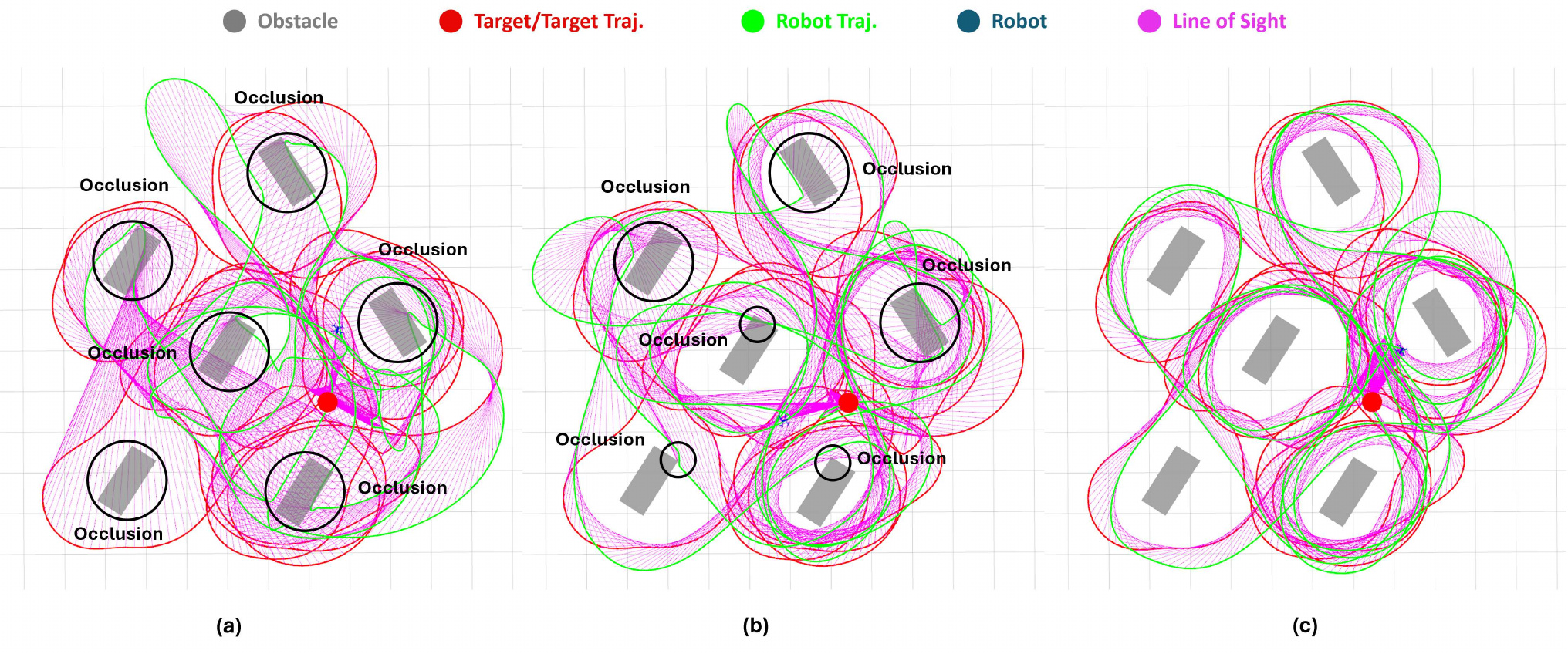}
\vspace{-0.4cm}
\caption{ (a) - (c) show the target tracking results for \textbf{Proj-CEM} \cite{ral_vis_aware}, conventional \textbf{BC}, and our proposed approach, respectively. Here the target is moving with the max speed of 2$m/s$. Both \textbf{Proj-CEM} and \textbf{BC} get occluded while going around the obstacle with the latter slightly outperforming the former. Our proposed method, however, perfectly follows the target and finishes the task without any occlusions.}
  \label{2m_comparison_figure}
  \vspace{-0.5cm}
\end{figure*}

\noindent Here, we increased the target speed to 2m/s. Given that AutoChaser \cite{auto_chaser_1} performed poorly in the previous section with target speed of 1/ms, we compare our approach only with the other two baselines in this benchmark. Fig. \ref{2m_comparison_figure} shows the qualitative results for this comparison. As the target speed increases, the performance of \textbf{Proj-CEM} \cite{ral_vis_aware} and \textbf{BC} degrades significantly. \textbf{Proj-CEM} \cite{ral_vis_aware} gets occluded around each obstacle while the \textbf{BC} performs better by having less occlusions around three of the obstacles. Our approach outperforms both the baselines and completes the task with no occlusions.

The second half of Table \ref{1m_comp_table} presents the quantitative results for these benchmarks. \textbf{Proj-CEM} \cite{ral_vis_aware} had the highest occlusion-time of approximately 62 seconds and the lowest success-rate in this benchmark. \textbf{BC} achieved better performance, having almost $40\%$ of the occlusion-time of \textbf{Proj-CEM} and consequently higher success-rate. Our approach performed the best with only 2.4 seconds of target occlusion along all the trajectory runs. All the approaches had approximately the same acceleration profiles.    

\begin{table}[h]
\centering
\caption{\scriptsize{Target Tracking in Static Environments}}
\scriptsize
\vspace{-0.2cm}
\begin{tabular}{|c|c|c|c|c|}
\hline
Method - Speed & \vtop{\hbox{\strut Occ. Time (s) /}\hbox{\strut Success-Rate}} & \vtop{\hbox{\strut Acceleration ($m/s^2$)}\hbox{\strut Mean / Min / Max}}  & \vtop{\hbox{\strut Comp.}\hbox{\strut Time (s)}} \\ \hline
\rowcolor{my_green}
Ours - $1m/s$ & 0 / 1 & 0.40 / 0 / 1.34 & 0.025 \\ \hline 
BC - $1m/s$ & 0.5 / 0.96 & 0.326 / 0 / 1.52 & 0.01 \\ \hline 
Ref \cite{ral_vis_aware} - $1m/s$ & 0.7 / 0.95 & 0.432 / 0 / 1.26 & 0.06 \\ \hline
\vtop{\hbox{\strut Ref. \cite{auto_chaser_1} - $1m/s$, no}\hbox{\strut prior target waypoints}} & 72.68 /0.0 & 0.312 / 0.00 / 1.14 & 0.085\\ \hline
\vtop{\hbox{\strut Ref. \cite{auto_chaser_1} - $1m/s$, 100}\hbox{\strut prior target waypoints}}  & 45.42 /0.10 & 0.478 / 0.00 / 1.35 & 0.085\\
\specialrule{2pt}{0pt}{0pt}
\rowcolor{my_green}

Ours - $2m/s$ & 2.4 / 0.92 & 0.48 / 0 / 1.82 & 0.025 \\ \hline
BC - $2m/s$& 38.2 / 0.74 & 0.41 / 0 / 1.7 & 0.01 \\ \hline
Ref \cite{ral_vis_aware} - $2m/s$  & 62.4 / 0.43 & 0.45 / 0 / 1.68 & 0.06 \\ \hline
\end{tabular}
\normalsize
\label{1m_comp_table}
\vspace{-0.7cm}
\end{table}

\subsection{Dynamic Obstacle Benchmark}

\noindent Here, we considered target tracking in dynamic environments with six moving obstacles. The robot needs to follow the target while avoiding collisions/occlusions stemming from the obstacles. Since AutoChaser \cite{auto_chaser_1} only works in static environment, we compared our approach against \textbf{BC} and \textbf{Proj-CEM} \cite{ral_vis_aware}. The obstacles and target are moving with speeds of $0.7m/s$ and $1.5m/s$, respectively. All the algorithms work without a prior map and they only had access to instantaneous LiDAR point-clouds. 

Table \ref{dynamic_comp_table} shows the quantitative results for this comparison. \textbf{Proj-CEM} \cite{ral_vis_aware} and \textbf{BC} achieved similar results with occlusion times of 40.2 and 44.5 seconds, respectively. Both these approaches also collided with obstacles during the movements. In sharp contrast, our approach had less than two seconds of occlusion throughout the whole trajectory run with zero collisions. All the methods used similar acceleration efforts. Due to lack of space, the qualitative results are presented in the accompanied video.

\begin{table}[h]
\centering
\caption{\scriptsize{Target Tracking in Dynamic Environments}}
\vspace{-0.2cm}
\scriptsize
\begin{tabular}{|c|c|c|c|c|}
\hline
Method - Speed & \vtop{\hbox{\strut Occ. Time (s) /}\hbox{\strut Success-Rate}} & \vtop{\hbox{\strut Acceleration ($m/s^2$)}\hbox{\strut Mean / Min / Max}}  & \vtop{\hbox{\strut Comp.}\hbox{\strut Time (s)}} \\ \hline
\rowcolor{my_green}
Ours - $1.5m/s$ & 1.4 / 0.94 & 0.48 / 0 / 1.85 & 0.025 \\ \hline
BC  - $1.5m/s$  & 44.5 / 0.63 & 0.41 / 0 / 1.72 & 0.01 \\ \hline
Ref \cite{ral_vis_aware}  - $1.5m/s$ & 40.2 / 0.68 & 0.45 / 0 / 1.78 & 0.06 \\ \hline
\end{tabular}
\normalsize
\label{dynamic_comp_table}
\vspace{-0.6cm}
\end{table}

\subsection{Ablation Study}\label{ablation}
\noindent In this section, we present an ablation of our approach, where the parameter $\mathbf{q}$ that spans the terminal state of the robot as well as the minimum and maximum tracking distances is not predicted by the learned model. Instead, we hard-code it to some nominal values. The qualitative result of this analysis is presented in the accompanying video. As it can be seen in Table \ref{ablation_table}, hand-specified $\mathbf{q}$ results in substantially worse performance as compared to our original approach where $\mathbf{q}$ is also predicted by the CVAE decoder. The performance degradation is more stark in the dynamic environments where it is more critical for the robot to adjust its separation from the target in order to improve obstacle avoidance performance.

\begin{table}[h]
\vspace{-0.2cm}
\centering
\caption{\scriptsize{Ablation in Static and Dynamic Environments}}
\vspace{-0.2cm}
\scriptsize
\begin{tabular}{|c|c|c|c|c|}
\hline
Method & \vtop{\hbox{\strut Occ. Time (s) /}\hbox{\strut Success-Rate}} & \vtop{\hbox{\strut Acceleration ($m/s^2$)}\hbox{\strut Mean / Min / Max}}  & \vtop{\hbox{\strut Comp.}\hbox{\strut Time (s)}} \\ \hline

\vtop{\hbox{\strut Ours, with $\mathbf{q}$ prediction}\hbox{\strut Static environment}}  & 1.4 / 0.94 & 0.48 / 0 / 1.85 & 0.025 \\ \hline
\vtop{\hbox{\strut Ours, no $\mathbf{q}$ prediction}\hbox{\strut Static environment}} & 10.3 /0.82 & 0.48 / 0 / 1.82 & 0.025 \\ \hline

\vtop{\hbox{\strut Ours, with $\mathbf{q}$ prediction}\hbox{\strut Dynamic environment}} & 2.4 / 0.92 & 0.48 / 0 / 1.82 & 0.025 \\ \hline
\vtop{\hbox{\strut Ours, no $\mathbf{q}$ prediction}\hbox{\strut Dynamic environment}} & 72.68 /0.0 & 0.312 / 0.00 / 1.14 & 0.025\\ \hline

\end{tabular}
\normalsize
\label{ablation_table}
\vspace{-0.5cm}
\end{table}

\subsection{Performance Under Resource Constraints}
\noindent Thanks to the differentiable optimization layer integrated into our CVAE, our method achieved good performance with a batch size as small as five. This small batch size enabled our method to operate with a computation time of 0.09 seconds on NVIDIA Jetson TX2. We observed that our approach was fast enough for maintaining occlusion-free runs when the maximum target speed was 0.5 m/s. For target speeds in the range of 1 m/s, the occlusion time was around 3.8 seconds out of the total run time of approximately 1020 seconds. As it can be seen from Table \ref{1m_comp_table}, this performance is better than what \textbf{Proj-CEM} \cite{ral_vis_aware} and  \textbf{AutoChaser} \cite{auto_chaser_1} achieved on a GPU RTX 3080 laptop.

\section{Conclusion}

\noindent This paper introduces a novel approach to address the challenges of tracking targets in cluttered and dynamic environments. By leveraging a learned probabilistic policy, which combines generative modeling with differentiable optimization layers, we demonstrated safe, agile and occlusion-free target tracking in challenging environments. Our approach surpasses the state-of-the-art in terms of occlusion/collision avoidance capabilities and computation time, as demonstrated through comparative analysis. Additionally, through extensive ablation studies, we provide insights into the contributions of different components within our learning pipeline. Moreover, the real-time performance of our approach on resource-constrained hardware underscores its practical feasibility. 
\section{Appendix} \label{append}
This section first presents a reformulation of constraints \eqref{uav_v_bounds}-\eqref{coll_avoid} and then show how they lead to the differentiable fixed-point iteration form presented in \eqref{fixed_point_form}.

\subsection{Reformulating Inequality Constraints}

\noindent We reformulate the quadratic inequality constraints \eqref{uav_v_bounds}-\eqref{coll_avoid} into \eqref{vel_reform}-\eqref{coll_avoid_reform}, respectively. The variables $(\alpha_v(t), \alpha_a(t), \alpha_r(t), \alpha_{d, i}(t) )$  and $(d_v(t), d_a(t), d_r(t), d_{ i}(t) )$ are additional variables that will be obtained with our optimizer along with $\boldsymbol{\xi}$.  As can be seen, our reformulated forms have a combination of non-convex equality and inequality constraints. Importantly, the inequality parts are just simple bounds.
\small
\begin{align}
    \dot{\mathbf{p}}(t) = d_v(t)\begin{bmatrix}
        \cos\alpha_v(t) & \sin\alpha_v(t)
    \end{bmatrix}^T\hspace{-0.2cm}, \hspace{-0.5cm}\qquad 0\leq d_v(t)\leq v_{max}
    \label{vel_reform}
\end{align}
\vspace{-0.6cm}
\begin{align}
    \ddot{\mathbf{p}}(t) = d_a(t)\begin{bmatrix}
        \cos\alpha_a(t) & \sin\alpha_a(t)
    \end{bmatrix}^T, \hspace{-0.5cm} \qquad 0\leq d_a(t)\leq a_{max}
    \label{acc_reform}
\end{align}
\vspace{-0.6cm}
\begin{align}
    {\mathbf{p}}(t)-\mathbf{p}_r(t) = d_r(t)\begin{bmatrix}
        \cos\alpha_r(t) & \sin\alpha_r(t)
    \end{bmatrix}^T, \nonumber \\ \underline{s}_{los}(t)\leq d_r(t)\leq \overline{s}_{los}(t)
    \label{track_reform}
\end{align}
\vspace{-0.7cm}
\begin{align}
    {\mathbf{p}}(t)-\mathbf{p}_{o, i}(t) = d_{o, i}(t)\begin{bmatrix}
        l\cos\alpha_{o, i}(t) & l\sin\alpha_{o, i}(t)
    \end{bmatrix}^T \nonumber \\, 0\leq d_{o, i}(t)\leq 1
    \label{coll_avoid_reform}
\end{align}
\normalsize

\noindent Using the parametrization introduced in \eqref{param}, we can put the equality part of \eqref{vel_reform}-\eqref{coll_avoid_reform} in the following compact form
\vspace{-0.2cm}
\begin{align}
    \mathbf{F}\boldsymbol{\xi}  = \mathbf{e}(\boldsymbol{\alpha}, \mathbf{d})
\end{align}
\vspace{-0.5cm}
\small
\begin{align}
   \mathbf{F} = \begin{bmatrix}
    \begin{bmatrix}
    \mathbf{F}_{o}\\
    \dot{\mathbf{W}}\\
    \ddot{\mathbf{W}}\\
    \mathbf{W}
    \end{bmatrix} & \mathbf{0}\\
    \mathbf{0} & \begin{bmatrix}
    \mathbf{F}_{o}\\
    \dot{\mathbf{W}}\\
    \ddot{\mathbf{W}}\\
    \mathbf{W}
    \end{bmatrix} 
    \end{bmatrix}, \mathbf{e} = \begin{bmatrix}
    \mathbf{x}_o+a \mathbf{d}_{o}\cos\boldsymbol{\alpha}_{o}\\
     \mathbf{d}_{v}\cos\boldsymbol{\alpha}_{v}\\
  \mathbf{d}_{a}\cos\boldsymbol{\alpha}_{a}\\
  \mathbf{x}_r +\mathbf{d}_{r}\cos\boldsymbol{\alpha}_{r} \\
 \mathbf{y}_o+a \mathbf{d}_{o}\sin\boldsymbol{\alpha}_{o}\\
     \mathbf{d}_{v}\sin\boldsymbol{\alpha}_{v}\\
  \mathbf{d}_{a}\sin\boldsymbol{\alpha}_{a}\\
  \mathbf{y}_r+\mathbf{d}_{r}\sin\boldsymbol{\alpha}_{r} \\
    \end{bmatrix}\\
    \boldsymbol{\alpha} = (\boldsymbol{\alpha}_{o}, \boldsymbol{\alpha}_{r}, \boldsymbol{\alpha}_{v}, \boldsymbol{\alpha}_{a})\\
    \qquad \mathbf{d} =  (\mathbf{d}_{o}, \mathbf{d}_{r}, \mathbf{d}_{v}, \mathbf{d}_{a})
    \label{W_matrix}
\end{align}
\normalsize

\noindent The matrix $\mathbf{F}_o$ is obtained by stacking the matrix $\mathbf{W}$ from (\ref{param}). Specifically, we repeat  $\mathbf{W}$ as many times as the number of point obstacles in LiDAR scans at a given planning cycle. The vectors $\mathbf{x}_o, \mathbf{y}_o, \mathbf{x}_r, \mathbf{y}_r$ are formed by appropriately stacking $x_{o, i}(t), y_{o, i}(t), x_r(t), y_r(t)$ at different time instants.  Similar construction is followed to obtain $\boldsymbol{\alpha}_{o},
\boldsymbol{\alpha}_{a}, \boldsymbol{\alpha}_{r}, \boldsymbol{\alpha}_{v}, \boldsymbol{\alpha}_{a}, \mathbf{d}_{o}, \mathbf{d}_{r}, \mathbf{d}_{v}, \mathbf{d}_{a}$. The vectors $\underline{\textbf{d}}, \overline{\textbf{d}}$ are constructed by stacking the lower and upper bounds of individual $d$'s from \eqref{vel_reform}-\eqref{coll_avoid_reform}. 

We now present a reformulation of the projection optimizer \eqref{cost_reform_2}-\eqref{eq_reform_2} embedded within the CVAE training pipelines as follows.

\vspace{-0.5cm}
\small
\begin{subequations}
\begin{align}
    \min_{{\boldsymbol{\xi}} }\frac{1}{2}\left\Vert {\boldsymbol{\xi}}-\overline{\boldsymbol{\xi}}\right\Vert_2^2 \label{cost_reform_3}  \\
    \textbf{A} {\boldsymbol{\xi}}= \textbf{b}(\mathbf{q}) \label{eq_reform_3} \\
    \textbf{F} {\boldsymbol{\xi}} = \textbf{e}(\boldsymbol{\alpha}, \textbf{d} ), \qquad \underline{\textbf{d}}(\mathbf{q})\leq \textbf{d}\leq \overline{\textbf{d}}(\mathbf{q}) \label{nonconvex_reform_3}
\end{align}
\end{subequations}
\normalsize

\subsection{Differentiable Solution Process} 
\noindent To solve, \eqref{cost_reform_3}-\eqref{nonconvex_reform_3}, we first relax the non-convex equality as augmented Lagrangian penalties and append them to the cost function in the following manner for some $\rho>0$.

\small
\begin{equation}
       \hspace{-1cm} \mathcal{L} (\boldsymbol{\xi}, \boldsymbol{\lambda}) = \frac{1}{2}\left\Vert \boldsymbol{\xi}-\overline{\boldsymbol{\xi}}\right\Vert_2^2-\langle \boldsymbol{\lambda}, {\boldsymbol{\xi}}\rangle\\+\frac{\rho}{2} \left \Vert \textbf{F} {\boldsymbol{\xi}}-\mathbf{e}(\boldsymbol{\alpha}, \textbf{d})\right \Vert_2^2,
    \label{aug_lag}
\end{equation}
\normalsize

\noindent where $\boldsymbol{\lambda}$ is the Lagrange multiplier which plays a crucial role in driving the residual of \eqref{nonconvex_reform_3} to zero. We minimize \eqref{aug_lag} through an Alternating Minimization routine presented in \eqref{am_alpha}-\eqref{am_xi}.

\vspace{-0.4cm}

\small
\begin{subequations}
    \begin{align}
        {^{k+1}\boldsymbol{\alpha}} &= \arg\min_{\boldsymbol{\alpha}} \mathcal{L}({^k}{\boldsymbol{\xi}}, {^k}\textbf{d}, \boldsymbol{\alpha}) = \mathbf{h}_1({^k}{\boldsymbol{\xi}}) \label{am_alpha}\\
        {^{k+1}\mathbf{d}} &= \arg\min_{\mathbf{d}} \mathcal{L}({^k}{\boldsymbol{\xi}}, \mathbf{d}, {^{k+1}}\boldsymbol{\alpha}({^k}{\boldsymbol{\xi}}) ) = \mathbf{h}_2({^k}{\boldsymbol{\xi}}) \label{am_d} \\ 
        {^{k+1}}\boldsymbol{\lambda} &= {^{k}}\boldsymbol{\lambda}+\rho \mathbf{F}^T (\mathbf{F}{^k}\boldsymbol{\xi}-\mathbf{e}( {^{k+1}}\boldsymbol{\alpha}, {^{k+1}}\mathbf{d}    )    ) \label{am_lambda} \\
        {^{k+1}}{\boldsymbol{\xi}} &= \arg\min_{\mathbf{A}\boldsymbol{\xi} = \mathbf{b}(\mathbf{q})}\mathcal{L}({\boldsymbol{\xi}}, {^{k+1}}\boldsymbol{\lambda}, {^{k+1}}\mathbf{e} ) \label{am_xi}\\
        &=\arg\min_{\mathbf{A}\boldsymbol{\xi} = \mathbf{b} (\mathbf{q})}\frac{1}{2}\left\Vert \boldsymbol{\xi}-\overline{\boldsymbol{\xi}}\right\Vert_2^2-\langle \boldsymbol{\lambda} \overline{\boldsymbol{\xi}}\rangle+\frac{\rho}{2} \left \Vert \textbf{F} {\boldsymbol{\xi}}-{^{k+1}}\mathbf{e}\right \Vert_2^2 \nonumber \\
        & = \mathbf{M}^{-1}\boldsymbol{\eta}({^k}\boldsymbol{\xi},~  {^{k}}\boldsymbol{\lambda}, ~\overline{\boldsymbol{\xi}},~ \mathbf{q}) \label{mvp}
    \end{align}
\end{subequations}
\normalsize
\vspace{-0.8cm}

\small
\begin{align}
    \textbf{M} = \begin{bmatrix}
        \mathbf{I}+\rho\mathbf{F}^T\mathbf{F} & \textbf{A}^{T} \\ 
        \mathbf{A} & \mathbf{0}
    \end{bmatrix}^{-1} \hspace{-0.19cm}, \boldsymbol{\eta} = \begin{bmatrix}
        \rho\mathbf{F}^T {^{k+1}}\mathbf{e}+{^{k+1}}\boldsymbol{\lambda}+\overline{\boldsymbol{\xi}}\\
        \mathbf{b}(\mathbf{q})
    \end{bmatrix} 
\end{align}
\normalsize

As can be seen, in each of the steps, we optimize only one set of variables while others are held fixed at the values obtained in the previous iteration or the preceding step of the current one. The optimization over $\boldsymbol{\alpha}$ in step \eqref{am_alpha} has a closed-form solution that depends only on ${^k}\boldsymbol{\xi}$. Similar solution structure is also obtained for optimization over $\mathbf{d}$ shown in \eqref{am_d} \cite{adajania2023amswarm}. In \eqref{am_lambda}, we update the Lagrange multiplier $\boldsymbol{\lambda}$. The optimization in step \eqref{am_xi} is an equality constrained Quadratic Programme and thus can be effectively reduced to a matrix vector product shown in \eqref{mvp}. Since, every step of our optimization routine involves a closed-form solution, it can be unrolled into a differentiable computational graph as shown in Fig. \ref{fig:proj_iteration}.

\subsection{Practical Considerations}

\noindent \textbf{Batched Operation over GPUs:} For efficient training, the projection optimizer should be capable of running in a batched fashion, wherein the projection of $\overline{\boldsymbol{\xi}}$ is parallelized across GPU cores. To this end, we point out that the functions $\mathbf{h}_1, \mathbf{h}_2$ have a symbolic form that does not require any matrix factorization or matrix-vector products. Thus, it can be trivially batched. Furthermore, the batched computation of Lagrange multiplier update in \eqref{am_lambda} and $\boldsymbol{\xi}$ update in \eqref{am_xi} can be formulated in terms of matrix-matrix products that can be trivially accelerated over GPUs. It should be noted that $\mathbf{M}$ is independent of $\mathbf{q}$ and the input $\overline{\boldsymbol{\xi}}$ that we are projecting to the feasible set. Thus, $\mathbf{M}$ remains the same across all the different instances/batches of the projection optimizer.

\noindent \textbf{Matrix Factorization Free Backpropgation:} Due to the computational structure of our projection optimizer, the matrix $\mathbf{M}$ does not contain any learnable parameters. Thus, its inverse/factorization needs to be computed only once before the training is started. This dramatically improves the speed and numerical stability of our training pipeline.  

\noindent \textbf{Optimizer Feasibility During Training:} A key practical challenge faced during training with differentiable optimization layers is that the optimization problem may become infeasible during training, at which point the whole back-propagation can get disrupted. This is particularly common when there are learanble parameters in the optimization layer itself, as in the case of our current approach. In this context, we point out that the optimization steps \eqref{am_alpha}-\eqref{mvp} is feasible for any arbitrary parameter $\mathbf{q}$ predicted by the CVAE decoder.

\bibliographystyle{IEEEtran}
\bibliography{paper}

\end{document}